\documentclass[journal,10pt]{IEEEtran}
\usepackage{makecell}
\usepackage{amsmath,amsfonts}
\usepackage{algorithmic}
\usepackage{algorithm}
\usepackage{array}
\usepackage[caption=false,font=normalsize,labelfont=sf,textfont=sf]{subfig}
\usepackage{textcomp}
\usepackage{stfloats}
\usepackage{url}
\usepackage{verbatim}
\usepackage{threeparttable}
\usepackage{graphicx}
\usepackage{cite}
\usepackage{booktabs}
\usepackage{multirow}
\usepackage{amssymb}
\usepackage{xcolor}
\usepackage{hyperref}
\usepackage{adjustbox}
\usepackage{bm}      

\newcommand{\revise}[1]{{\color{black}#1}}

\hypersetup{
    colorlinks=true,
    linkcolor=blue,
    filecolor=magenta,      
    urlcolor=cyan,
    pdfpagemode=FullScreen,
    }
\newtheorem{definition}{Definition}

\ifCLASSINFOpdf
\else
\fi
\begin{document}
\begin{sloppypar}
%
\title{Relation-Aware Distribution Representation Network for Person Clustering with Multiple Modalities}
%
%
%

\author{Kaijian Liu, Shixiang Tang, Ziyue Li$^*$, Zhishuai Li, Lei Bai, Feng Zhu, and Rui Zhao
\thanks{Kaijian Liu, Zhishuai Li, Feng Zhu, Rui Zhao are with the SenseTime Research, 200030, Shanghai, China}
\thanks{Shixiang Tang is with The University of Sydney, Sydney, NSW, Australia.}
\thanks{Ziyue Li is with the University of Cologne, 50923, Cologne, Germany. He is also with the EWI gGmbH, 50827, Cologne, Germany.}
\thanks{Lei Bai is with the Shanghai AI Laboratory, 200030, Shanghai, China.}
\thanks{Rui Zhao is also with the Qing Yuan Research Institute of Shanghai Jiao Tong University, 200040, Shanghai, China.}
\thanks{*Corresponding Author. Email: \url{zlibn@wiso.uni-koeln.de}}
}

%
%

\markboth{Accepted in IEEE Transactions on Multimedia, August~2023}%
{Liu \MakeLowercase{\textit{et al.}}: Relation-Aware Distribution Network for Multi-Modality Person Clustering}
%



\maketitle

\begin{abstract}
Person clustering with multi-modal clues, including faces, bodies, and voices, is critical for various tasks, such as movie parsing and identity-based movie editing. Related methods such as multi-view clustering mainly project multi-modal features into a joint feature space. However, multi-modal clue features are usually rather weakly correlated due to the semantic gap from the modality-specific uniqueness. As a result, these methods are not suitable for person clustering.
In this paper, we propose a \textbf{R}elation-\textbf{A}ware \textbf{D}istribution representation Network (RAD-Net) to generate a \textit{distribution representation} for multi-modal clues.  The distribution representation of a clue is a vector consisting of the relation between this clue and all other clues from all modalities, thus being \textit{modality agnostic} and good for person clustering. Accordingly, we introduce a graph-based method to construct distribution representation and employ a cyclic update policy to refine distribution representation progressively. Our method achieves substantial improvements of \textbf{+6\%} and \textbf{+8.2\%} in F-score on the Video Person-Clustering Dataset (VPCD) and VoxCeleb2 multi-view clustering dataset, respectively. Codes will be released publicly upon acceptance.
\end{abstract}

\begin{IEEEkeywords}
Person clustering, Multi-modality clues, Distribution learning, Multi-modal representations
\end{IEEEkeywords}

\title{Relation-Aware Distribution Representation Network for Person Clustering with Multiple Modalities}
\maketitle

\section{Introduction}
\IEEEPARstart{U}{nderstanding} videos~\cite{wang2020story,siddique2021object} such as TV series and movies has been a prior step to various vision tasks such as story understanding~\cite{zhu2015aligning,lavee2009understanding,wu2019long}, browsing movie collections~\cite{ansari2015content,low2017exploring}, and identity-based video editing~\cite{wang2020story,siddique2021object}. However, it relies on identifying the characters and analyzing behaviors, considering characters are always core elements of any story. Characters in videos are often presented in the form of the person tracks~\cite{kalogeiton2020constrained,brown2021face}, which are video clips including face, body, and voice information. Before further analysis~\cite{kim2020modality,huang2020location,yang2020bert}, 
a vital research subject is the $\emph{identification}$~\cite{deng2019arcface,zheng2021online,chen2016similarity,chen2015similarity}, which requires labeling person tracks based on their identities. Therefore, a person clustering task is proposed by~\cite{brown2021face} to cluster a large number of person tracks in videos based on their identities by considering a person's multi-modal clues, \emph{e.g.,} face images, body images, and voices. 

\begin{figure}[t]
\centering
\includegraphics[width=0.99\linewidth]{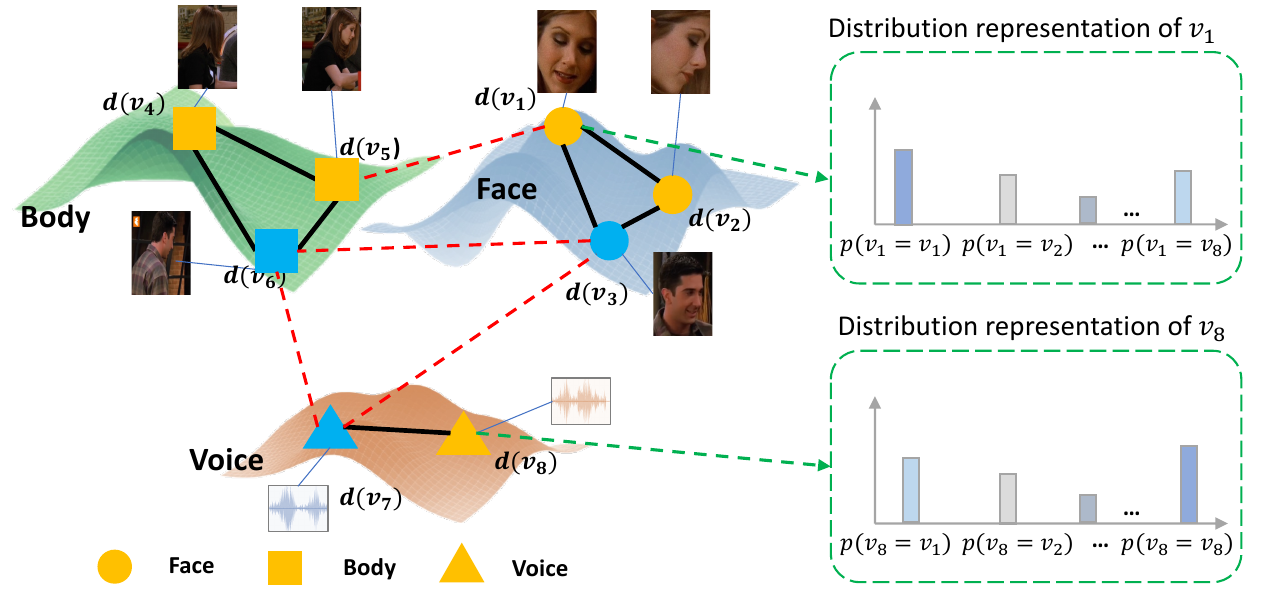} 
\caption{\small{Illustration of distribution representation for person clustering. Orange and blue indicate different identities. Circles, squares, and triangles indicate different modalities. The distribution representations are generated by inference on a probability graph, where the black solid lines denote intra-modality edges, and the red dotted lines denote inter-modality edges. Distribution representations of $v_1$ and $v_8$ are generated by the relation among all clues. }
}
\label{fig:teaser_motivation}
\end{figure}

Unlike the well-developed face clustering~\cite{wang2019linkage,guo2020density,yang2020learning}, the person clustering task is more challenging because it requires dealing with multi-modal clues rather than one single modal clue. For person clustering, clue features of different modalities are extracted by various feature extractors: features only contain modality-specific information, which brings significant \textit{semantic gap} between different modalities, as shown in Fig. \ref{fig:teaser_motivation}. 

To the best of our knowledge, there are no specific methods to cluster person, yet a few methods were proposed to tackle multi-modal representations, such as multi-view clustering~\cite{rupnik2010multi,zhang2019ae2,ma2021discriminative,lin2021completer}. These methods mainly follow the guideline to project multi-modal features into a common space where multi-modal features are statistically correlated. However, it is rather challenging to project all different modality features, \emph{i.e.,} faces, bodies, and voices, into a joint space since there is a large semantic gap: For example, the face feature is invariant to color~\cite{wang2018cosface,meng2021magface}, whereas the body feature is easily affected by cloth color~\cite{chen2016similarity,ge2020mutual,zheng2021online}; Both face and body features are sensitive to light condition, but voice feature isn't; Lastly, a person could have multiple face and body features, but usually only one voice feature. Thus, due to the weak correlation between different modalities, it is difficult to project different modality features into a joint space to get the relationship between two samples of different modalities. As a result, existing multi-view clustering achieves low performance on the person clustering.

In this paper, our solution comes outside the box: we no longer constrain ourselves within the idea of ``projection to a common space''. Instead, we look at the person's multi-modal clustering from a new perspective: \textit{distribution representation}. We define the distribution representation as a vector containing the \textit{relations} of a cue with all the other cues. The \textit{relation} is defined as the probability of two cues coming from the same person. 

It is worth emphasizing the difference between our distribution representation and traditional feature representation: the extracted clue features are strictly modality-specific, whereas distribution aims to preserve relations with other clues rather than keeping modality-specific vision or audio information, so that it will not be affected by the semantic gap of different modalities. Thus, distribution representations are constructed based on the relations across \textbf{all} clues, being irrelevant to specific modal, \emph{i.e.,} agnostic to modalities. As a result, if two cues are from the same identity, their distribution vectors are designed to be similar regardless of whether they are from the same modality. Owing to this good property, distribution representations can be directly used to easily cluster the same identities together, even with different modalities.

To this end, we design a \textbf{R}elation-\textbf{A}ware \textbf{D}istribution representation Network (\textbf{RAD-Net}) to cluster multi-modal person clues. The framework of RAD-Net is summarized as follows: Firstly, we establish the relation-aware distribution representation from clue features by probabilistic inferences, followed by a momentum update mechanism to get a more precise distribution representation. Secondly, the distribution similarity is employed to enhance feature representation via a cyclic update policy, such that distribution and feature representation could contribute to each other. With the robust distribution representations refined cyclically, we can cluster multi-modal clues directly. 

In summary, our contributions are three-fold. 
\begin{itemize}
    \item \revise{We propose a relation-aware distribution representation, which contains global and impartial information from all modalities, thus being modality-agnostic.} This  distribution representation can directly measure identity-based similarities between multi-modal clues for person clustering.
    \item We introduce a graph-based method to establish distribution representation. \revise{This distribution representation could further improve the feature representation by indicating how to aggregate the multiple features, achieving a cyclic updating between the distribution and the feature.} 
    \item We conduct intensive experiments comparing with the state-of-the-art methods. Our model achieves  \textbf{+6\%} and \textbf{+8.2\%} F-score on the large-scale Video Person-Clustering Dataset and VoxCeleb2 multi-view clustering dataset, respectively.
\end{itemize}

The rest of the paper is organized as follows. Section \ref{sec: related work} provides the literature review on related multi-model clustering and distribution learning. Section \ref{sec: preliminary} states person clustering formulation and definitions. Section \ref{sec: method} details the proposed RAD-Net. Numerical experiments are performed in Section \ref{sec: experiment}. The concluding remarks and future directions are discussed in Section \ref{sec: conclusion}.

\section{Related Work}
\label{sec: related work}

This section will introduce related works from two perspectives: (1) similar tasks, including face clustering and video hyperlinking, as well as (2) related techniques, including traditional distance-based learning, multi-modality and multi-view clustering, and distribution learning, respectively.

\subsection{Face Clustering}
Face clustering focuses on the single-modal features \textit{i.e.}, face features, yet person clustering tries to cluster multi-modal features together, such as face, body, and voice. 
Existing face clustering methods are divided into unsupervised and supervised clustering methods. Unsupervised face clustering methods focused on designing effective similarity metrics by considering the context~\cite{zhu2011rank,otto2017clustering,lin2017proximity,lin2018deep} in the feature space. Supervised methods tackled the problem by learning the metric with graph pooling~\cite{yang2019learning}, Transformer~\cite{ye2021learning}, or Graph Convolutional Network (GCN)~\cite{wang2019linkage,yang2020learning,guo2020density}. However, they can not be applied to person clustering because the feature similarities of multi-modal clues can not measure the identity-based similarities. 
In response to this concern, our RAD-Net constructs graphs in the distribution space where the similarity of two clues from different modalities can be directly computed by distribution representation similarities. Thus, we extend the graph-based methods from single-modal clustering to multi-modal clustering.

\subsection{\textcolor{black}{Video Hyperlinking}}
\revise{Another similar task is \textit{video hyperlinking}~\cite{awad2016trecvid, awad2017trecvid, cheng2017selection},  which is popular with the rise of video platforms such as YouTube and short video streaming such as TikTok. With the objective of improving the accessibility of vast video datasets, video hyperlinking establishes links between segments from various relevant videos, enabling users to seamlessly navigate between videos by utilizing hyperlinks. 

To link the anchors (source videos) and targets (destination videos), several technical approaches are proposed. \cite{eskevich2013multimedia} and Video-to-Text (VTT) \cite{nguyen2017vireo} proposed to link two videos by both visual clues and text clues, with ResNet-152 features extracted from frames and text features encoded by
LSTM. Ad-Hoc Video Search (AVS) further combines VTT and a text-based module that extracts the on-screen text ad speech text to achieve video-text search. Video Hyperlinking (LNK) \cite{nguyen2017vireo} further considered the multi-modal similarity of visual-visual, visual-text, text-visual, and text-text. As observed, though also claimed as multi-modal clustering, video hyperlinking only handles each video segment based on visual and text clues, and it is indiscriminately dealing with face, body, and voice clues, which are essential for a human-centric task like person clustering. 

This video hyperlinking has also influenced e-commerce. For example, video eCommerce++ \cite{cheng2017video} aims to exhibit appropriate product ads to particular uses at proper time stamps of video. Video eCommerce++ proposed to learn the video-product association via object detection on the sequence of keyframes. Together with a user-produce association, recommendations could be directed to specific users. However, this framework only utilized the object clues in the video for product association, thus not applicable to person clustering based on face, body, and voice clues.} 


\subsection{\revise{Distance Metric Learning}}
\revise{Distance metric learning is to learn a distance for various tasks, such as image retrieval~\cite{li2018deep, li2015weakly} and feature selection~\cite{li2015unsupervised}. Similarly, the person clustering task also relies on pairwise distance. However, Previous works mainly designed for single-modal metric learning, and they can not be applied to measure the distance between samples of different modalities. DCE~\cite{li2018deep} can be used for cross-modality retrieval by projecting images and text into a unified space. But We can not project faces/bodies/voices  features to a unified feature space since they are too weakly correlated, so projecting semantically different faces/bodies/voices into a unified space will have bad representations.}

\subsection{Multi-modality and Multi-view Clustering}
To the best of our knowledge, there are only a few methods of clustering a person using multi-modality~\cite{brown2021face,face_audio}. For example, Brown \textit{et al.} \cite{brown2021face} clustered clues within each modality first, then relied on manually-designed rules to fuse the multiple clues, which is not end-to-end trainable.

Multi-view clustering~\cite{chen2020multi,jing2021learning,wu2020unified,liang2020multi} is highly related to multi-modality clustering by treating different modalities as different views. However, the biggest difference is that multi-view clustering is typically for multi-view features extracted from the same instance, whereas person clustering is for clustering clues from both different modalities and different instances. Multi-view clustering methods mainly consider the diversity and complementarity of different views~\cite{andrew2013deep,xia2014robust,tang2018learning,tang2020cgd}, and try to project features from multiple modalities into one unified joint space~\cite{liu2017principled,zhang2018generalized,zhang2020tensorized,lin2021completer,zhou2020end}. 
For instance, MvSCN~\cite{huang2019multi} proposed a multi-view spectral clustering network to project multi-view features into a joint space by incorporating the local manifold invariance across different views. CONAN~\cite{ke2021conan} employs an encoder network to obtain view-specific features and a fusion network to get common representations. 
The above methods rely on strong cross-view feature correlation, which is not usually satisfied in person clustering due to semantic gaps. RAD-Net does not suffer from this problem because the distribution representation is generated by relations of samples, without relying on consistency across different modalities.

\subsection{Distribution Learning}
Distribution learning~\cite{kearns1994learnability} proposed to learn the distribution from which the samples are drawn, 
could also be used for classification tasks~\cite{kalai2010efficiently,yin2012facial,daskalakis2015learning,gao2017deep,shen2017label}. The most related work is DPGN~\cite{yang2020dpgn}. However, DPGN is designed for few-shot learning with single-modal samples, which fails to fuse information from multiple modalities. We design a probabilistic graph to utilize intra-modality similarity and inter-modality association, enabling RAD-Net to transform multi-modal information to the distribution space.

\section{Problem Definition}
\label{sec: preliminary}

\begin{figure*}[t]
\centering
\includegraphics[width=0.8\linewidth]{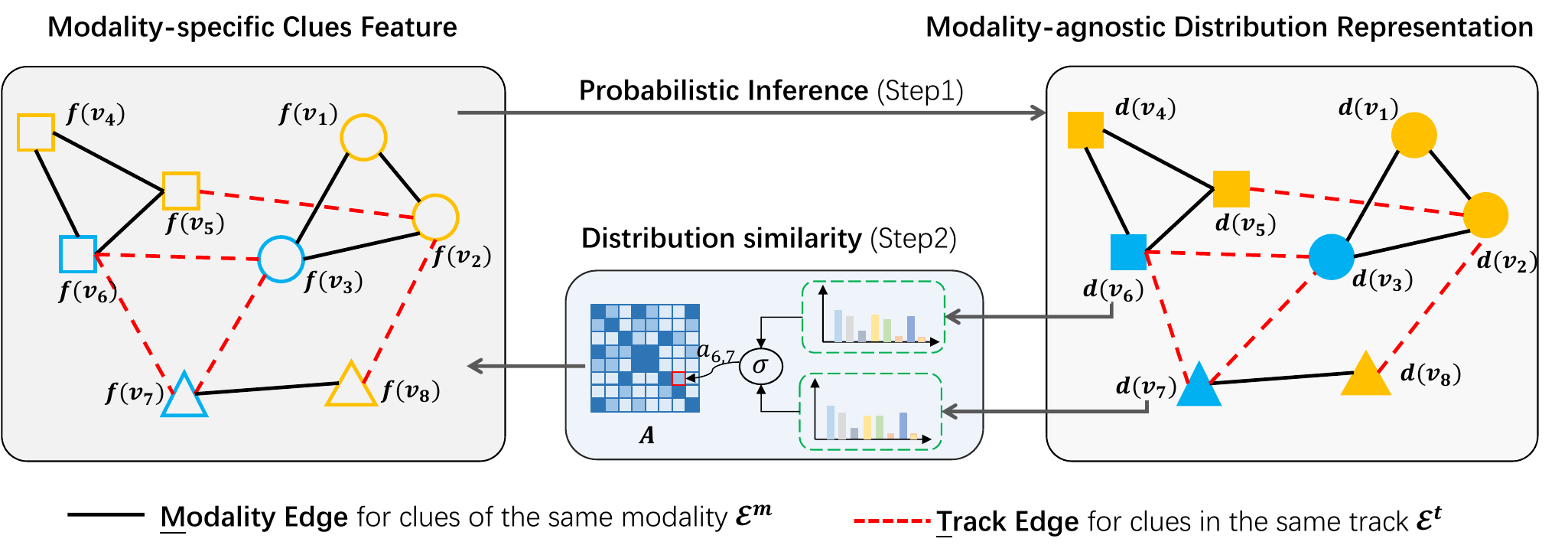} 
\caption{\small{The overall framework of the proposed method. Circle, square, and triangle denote face, body, and voice respectively, and different colors indicate different identities. Our method includes a cyclic update process: calculate distribution representation and update clue features. 
}}
\label{fig: framework}
\end{figure*}

\textit{\textbf{Problem Statement:}} Given a dataset $\mathbb{X} = \{\mathcal{X}_1, \mathcal{X}_2, .., \mathcal{X}_T\}$ with $T$ person-tracks, where $\mathcal{X}_i$ is $i$-th person-track, the goal is to cluster person tracks in $\mathbb{X}$ into $C$ clusters ($C$ is unknown). We denote the $i$-th person-track as $\mathcal{X}^i=\{\mathcal{F}^i, \mathcal{B}^i, \mathcal{U}^i\}$, where $\mathcal{F}^i$, $\mathcal{B}^i$, $\mathcal{U}^i$ are the face, body, voice modals, respectively. The availability of each modality feature depends on whether the face or body is visible or if they are speaking. Usually, there are multiple features in each face-modal and body-modal, but there is only one feature for the voice-modal~\cite{brown2021face}. We define $p$ face clues, $p$ body clues, and $q$ voice clues sampled from $q$ person tracks. 
\begin{definition}
\textbf{Multi-modal Graph} is defined as the graph of all multi-modal clues, \textit{i.e.},  $\mathcal{G} = (\mathcal{V}; \mathcal{E}^m, \mathcal{E}^t)$, where $\mathcal{V}\!=\{v_1, v_2, ..., v_N\}$ is the set of all sampled clues, and $N\!=\!2p+q$.
The \textbf{modality edge} $\mathcal{E}^m$ is defined as the edges between clue features of the same modality, shown as solid lines in Fig.~\ref{fig: framework}: $\mathcal{E}^m\!=\!\{e_{ij}|m(i)\!=\!m(j)\}$, where $m(i)$ is the modality of clue $v_i$.  The \textbf{track edge} $\mathcal{E}^t$ is defined as the edges between clue features in the \emph{same} track but of \emph{different} modalities, shown as dotted lines in Fig.~\ref{fig: framework}: $\mathcal{E}^t\!=\!\{e_{ij}|t(i)\!=\!t(j), m(i)\! \ne \!m(j)\}$, where $t(i)$ is the track ID of clue $v_i$. Since this graph could be quite large, Appendix I.A in supplementary material shows how to obtain a fixed-size graph via data sampling.
\end{definition}

\begin{definition}
\label{def: p}
\textbf{Identity Probability:} $p(v_i\!=\!v_j)$ represents the probability that $v_i$ and $v_j$ are of the same identity, \textit{i.e.}, $p(v_i\!=\!v_j)\!=\!\mathbb{P}[I(v_i)\!=\!I(v_j)]$, where $I(v_i)$, $I(v_j)$ are the identities of $v_i$ and $v_j$.
\end{definition}

\begin{definition}
\label{def: d}
\textbf{Distribution Representation:} \revise{As mentioned before, the distribution representation is a vector containing
the relations of a cue with all the other cues, and the relation is
defined as the probability of two cues coming from the same
person.} Specifically,  the distribution representation $\boldsymbol{d}(v_i)$ of $v_i$ contains \textit{identity probabilities} (in Definition \ref{def: p}) with \textbf{all} the clues regardless of modalities. $\boldsymbol{d}(v_i)= \left[ p(v_i\!=\!v_1), \dots, p(v_i\!=\!v_j), \dots, p(v_i\!=\!v_N) \right]$, $\boldsymbol{d}(v_i) \in \mathbb{R}^N$. \revise{The distribution representation is each clue-specific, and its entry value only relies on the relation between two clues.}
\end{definition}

\section{Methodology}
\label{sec: method}


For person clustering, clue features are extracted by corresponding feature extractors with modality-specific information, bringing the gap between clues of different modalities. To tackle this, we propose a modality-agnostic distribution representation. Unlike the traditional feature representation, distribution representation is constructed based on the novel multi-modal graph that collects the relations with all clues. \revise{This section is organized as follows: In Sec \ref{sec: model-overview} we give the overview. The \textit{distribution} is then defined as the concatenation of $1$-\textit{vs}-$n$ identity probabilities with all clues regardless of their modalities, \textit{a.k.a}, irrelevant to modality. Thus, we can directly cluster clues of different modalities, so in Sec \ref{sec:distribution_representation}, we propose to model clues of different modalities with distribution representation. The distribution representation relies on the pairwise similarity of clues. In Sec \ref{sec:distribution_representation},  distribution representation can guide how to aggregate the neighboring features to enhance the feature representation quality.}
\subsection{Framework Overview}\label{sec: model-overview}
The overall framework is shown in Fig. \ref{fig: framework}. 
Our proposed RAD-Net leverages a well-designed cyclic update strategy between clue features 
and distribution representations 
for $L$ cycles. 
For $l$-th cycle, we denote the clue features of $v_i$ as $\bm{f}^l(v_i) \in \mathbb{R}^{O}$ ($O$ as the dimension of clue feature), which is initialized by clue features available in the dataset $\bm{f}^0(v_i)$. Distribution representation based on multi-modal $\mathcal{G}$ is defined as $\bm{d}^l(v_i) \in \mathbb{R}^{N}$, with a soft initialization~\cite{tang2021mutual}:
\begin{equation} 
\bm{d}^0(v_i)_j=\left\{\begin{array}{cc}
\eta, & \mbox{if } t(i)=t(j),\\
1-\eta, & \mbox{if } t(i)\not=t(j),
\end{array}\right.
\label{eq_d_init}
\end{equation}
where $t(i)$ return the track ID of clue $v_i$ and $\eta \in [0,1]$ is a hyperparameter.
The pipeline of our proposed RAD-Net can be summarized in four steps:

\emph{\textbf{Step 1}: Calculating distribution representation $\bm{d}^{l+1}(v_i)$ with clue features $\bm{f}^l(v_i)$ (Sec.~\ref{sec:distribution_representation}).} Given the multi-modal graph $\mathcal{G}$ and the clue features $\bm{f}^l(v_i)$, we compute the distribution representation of $\bm{d}^{l+1}(v_i)$ by probabilistic inference. 

\emph{\textbf{Step 2}: Update the clue features $\bm{f}^{l+1}(v_i)$ by distribution representations $\bm{d}^{l+1}(v_i)$ (Sec.~\ref{sec:clue_features_update}).} Given the multi-modal graph $\mathcal{G}$ and $\bm{d}^{l+1}(v_i)$, we compute the clue features of $\bm{f}^{l+1}(v_i)$ by feature aggregation. 

\emph{\textbf{Step 3}: Cyclic update the clue features and distribution representations for $L$ cycles and optimize the network by backward propagation (Sec.~\ref{sec:objective}).}

\emph{\textbf{Step 4}: Clustering person tracks with the distribution representations after network optimization (Sec.~\ref{sec:final_cluster}).}


\subsection{Relation-Aware Distribution Representation} \label{sec:distribution_representation}
Different from the features which are specific to their modality, \revise{we aim to learn a distribution representation for each cue that is independent of its modality. We believe this modality-agnostic distribution representation offers more general and global information about which person cluster this cue belongs to. For instance, when two cues' distribution representations are quite similar, they are highly likely from the same person cluster.} 

Specifically, in the $l$-th cycle, we design a novel distribution representation $\bm{d}^l(v_i)$ of $v_i$ based on multi-modal graph $\mathcal{G}$, as shown in Definition~\ref{def: d}, and there is:
\begin{equation} 
   \bm{d}^l(v_i)= \left[ p^l(v_i\!=\!v_1), \dots, p^l(v_i\!=\!v_j), \dots, p^l(v_i\!=\!v_N) \right] ^ \top, 
\end{equation}
The key is to calculate identity probabilities in the $l$-th cycle.

\noindent \textbf{(1) When $v_i$ and $v_j$ are from the same modality:} denoted as an edge $e_{ij}$ existing in $\mathcal{E}^m$: $e_{ij} \in \mathcal{E}^m$, 
the relation is:
\begin{equation} 
p^l(v_i=v_j)=	\mathbb{P}[I(v_i)\!=\!I(v_j)] = \langle \bm{f}^{l-1}(v_i),\bm{f}^{l-1}(v_j) \rangle,
\label{eq_pij}
\end{equation}
where inner-product $\langle \bm{f}^{l-1}(v_i),\bm{f}^{l-1}(v_j) \rangle$ provides the similarity between the nodes $v_i$ and $v_j$, such as cosine similarity. 

\noindent \textbf{(2) When $v_i$ and $v_j$ are from different modalities:} \textit{i.e.}, $e_{ij} \notin \mathcal{E}^m$, there exist two situations: \textit{a)} $v_i$ and $v_j$ are from the same track, \textit{i.e.}, $e_{ij}\!\in\!\mathcal{E}^t$, or \textit{b)} different tracks $e_{ij}\!\notin\!\mathcal{E}^t$.

\underline{\textit{a)} \emph{If $v_i$ and $v_j$ are from the same track,}} the identity probability is defined as 1, \emph{i.e.,}
\begin{equation} 
    p^l(v_i=v_j) = 1,
\end{equation}

\underline{\emph{b) If $v_i$ and $v_j$ are from different tracks,}} we use a clue $v_k$ that has the same modality of $v_i$ ($e_{ik} \in \mathcal{E}^m$) and shares the same track ID of $v_j$ ($e_{kj}\in \mathcal{E}^t$) as a bridge: $v_k \in \{v_k|e_{ik} \in \mathcal{E}^m, e_{kj}\in \mathcal{E}^t\}$. 
An example in Fig. \ref{fig: framework} is: from $v_1$ to $v_7$, the bridge could be $v_3$. The identity probability is derived as:
\begin{equation}
  \scriptsize 
  \begin{aligned}
   p^l(v_i\!=\!v_j) &= \mathbb{P}[I(v_i)\!=\!I(v_j)] \\
          &= \sum_{v_k} \mathbb{P}[I(v_i)\!=\!I(v_k),I(v_k)\!=\!I(v_j)] \\
         & = \sum_{v_k}\mathbb{P}[I(v_i)\!=\!I(v_k)|I(v_k)\!=\!I(v_j)] \cdot \mathbb{P}[I(v_k)=I(v_j)]\\
         & = \sum\limits_{v_k}{p^l(v_i\!=\!v_k) p^l(v_k\!=\!v_j)}=\sum\limits_{v_k} p^l(v_i\!=\!v_k), \\
  \end{aligned}
\label{eq_pij_dif}
\end{equation}
where $p^l(v_k\!=\!v_j)\!=\!1$ due to $e_{kj}\in \mathcal{E}^t$, and $p^l(v_i\!=\!v_k)$ is given by Eq. (\ref{eq_pij}) due to $e_{ik} \in \mathcal{E}^m$. Normalization is applied so that $p^l(v_i\!=\!v_j) \in [0, 1]$. As a result, the identity probability of $v_i$ and $v_j$ is the summation of identity probabilities of all the intermediate nodes $v_k$ that bridge $v_i$ and $v_j$. 

To stabilize training, we update the distribution in a momentum way, \emph{i.e.}, $\bm{d}^l(v_i)\leftarrow\alpha \bm{d}^{l-1}(v_i)+(1-\alpha)\bm{d}^{l}(v_i)$, where $\alpha$ weighs the contribution of the two components.

\subsection{Modality-agnostic Distribution Enhances Modality-specific Feature}\label{sec:clue_features_update}

\revise{Once the modality-agnostic distribution representation is learned by using  features' pairwise relations, this global distribution vector could, in return, enhance the original modality-specific feature by aggregating the multiple features from the same person. The question is: which feature from the same modality is more likely to be from the same person? The proposed distribution representation offers the optimal solution: the global distribution has more impartial and accurate information about the person cluster since it leverages complementary information between different modalities and considers all clues in the graphs, so using the similarity of distribution representation to guide the aggregation is more precise than common options such as using feature similarity}. 

We enhance features to get a more robust feature representation \revise{by aggregating the multiple neighboring features from the same person in a weighted manner}. Specifically, we define the distribution similarity as $\mathbf{A}$, denoting the pairwise final probability of any two clues being from the same person $\mathbf{A}^l=\{a^l_{ij}|i,j \in \mathbb{R}^N\}$. Mathematically it can be computed as follows:
\begin{equation}
\small
a^l_{ij}=\sigma_l(|\bm{d}^l(v_i)-\bm{d}^l(v_j)|),
\label{eq_d_similarity}
\end{equation}
where $\sigma(\cdot)$: $\mathbb{R}^N\rightarrow{\mathbb{R}^1}$ gets the distribution similarity with two fully-connected layers and a sigmoid layer.

\begin{algorithm}[t]
\caption{\revise{Density-Aware Feature Sampling}}
\label{alg_sampling}
\textbf{Input}: samples $X$ from a track, total number of samples $\mathcal{M}$ of the track, parameters $q$\\
\textbf{Output}: Sampled features $\mathcal{S}$
\begin{algorithmic}[1] 
\FOR{$i$=1 to $\mathcal{M}$} 
\STATE Calculate the local density for $x_i$ by $\rho_i=\sum_{j=1}^{M}\delta(x_i,x_j)d(x_i,x_j)$
\STATE In the set of all the local density greater than $x_i$, the sample with the highest similarity to the $x_i$ is taken, and get the proximity density peak distance $r_i=1-d(x_i,x_j)$
\ENDFOR
\STATE Normalize $\rho_i$ and $r_i$ for all features, and get $\bar{\rho_i}$, $\bar{r_i}$
\STATE Obtain ranking score by $score_i=\bar{\rho_i}\bar{r_i}$
\STATE Construct $\mathcal{S}$ with $q$ samples with highest score from $X$
\STATE \textbf{return} $\mathcal{S}$
\end{algorithmic}
\end{algorithm}

With the distribution similarity matrix $\mathbf{A}^l$, the clue features $\bm{f}^l(v_i)$ are enhanced by the neighborhood aggregation with clue features of the \emph{same} modality, which can be formulated as follows:
\begin{equation}
    \bm{f}^{l+1}(v_i)=\phi_{l,m(i)}(\sum\limits_{e_{ij}\in \mathcal{E}^m}a_{ij}^l\cdot \bm{f}^{l}(v_i)),
    \label{eq_f_aggregation}
\end{equation}
where $\phi(\cdot):\mathbb{R}^O\rightarrow\mathbb{R}^O$ is the learnable gated residual block~\cite{shi2020masked}, which is modality-dependent.

\revise{As shown in the Sec. \ref{sec:distribution_representation} and \ref{sec:clue_features_update}, visualized in Fig. \ref{fig: framework}, the modality-specific features help to construct the modality-agnostic distribution representation, and the distribution representation further enhances the feature quality back: this cyclic mechanism encourages to learn both of the feature and distribution better.}

\subsection{Objective Function} \label{sec:objective}
We supervise the feature loss $\mathcal{L}_f$ and distribution loss $\mathcal{L}_d$ simultaneously to optimize $\sigma_l$ and $\phi_{l,m(i)}$ in Eq. (\ref{eq_d_similarity}) - (\ref{eq_f_aggregation}). The two loss functions are defined with the Binary Cross-Entropy (BCE) function as follows:
\begin{equation}
    \small
    \label{eq_f_loss}
    \mathcal{L}_f = \sum_{l=1}^L\sum_{i=1}^N\sum_{j=i+1}^N\mathbb{I}[m(i)=m(j)]\mu_l^f \mathbf{BCE}\left(y_{ij}, \langle \bm{f}^l(v_i),\bm{f}^l(v_j) \rangle\right),
\end{equation}
\begin{equation}
    \small
    \label{eq_d_loss}
    \mathcal{L}_d = \sum_{l=1}^L\sum_{i=1}^N\sum_{j=i+1}^N\mu_l^d\mathbf{BCE}(y_{ij}, a_{ij}^l).
\end{equation}
where  $\mathbf{BCE}(\cdot)$ denotes the BCE loss, and $\mu_l^f$ and $\mu_l^d\!$ are the weights for feature loss and distribution loss, respectively. $\mathbb{I}(x)$ is an indicator function. 
$y_{ij}=1$ if node $i$ and $j$ have the same label, otherwise $y_{ij}=0$. 

The final loss is defined as a weighted summation of two losses, with hyper-parameters $\lambda_p$ and $\lambda_d$:
\begin{equation}
    \mathcal{L} = \lambda_f\mathcal{L}_f+\lambda_d\mathcal{L}_d.
\end{equation}

\subsection{Clustering Tracks by Distribution} \label{sec:final_cluster}
With the similarity matrix $\mathbf{A}^L$ generated by distribution representation, the track-track linkage score can be obtained by the average pairwise similarity between all clues of the two tracks. We connect two tracks if their linkage is higher than a threshold, and we group connected tracks into one cluster by Union-Find algorithm~\cite{fredman1989cell}.

\begin{figure}[t]
\centering
\includegraphics[width=0.26\textwidth]{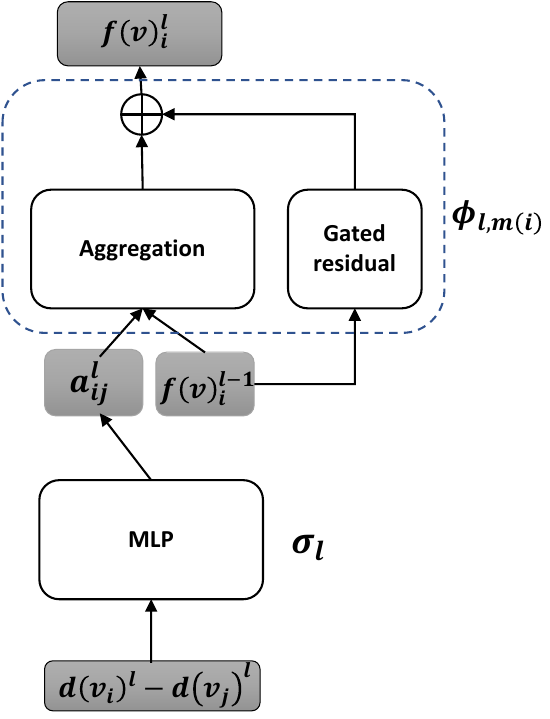} 
\caption{\revise{Detailed network architectures used in RAD-Net.}}

\label{fig_network}
\end{figure}

\subsection{Model Details}
\revise{
\subsubsection{Data Sampling}
Generally, there are many features in a track, which will bring redundancy and massive computation if all the features of a track are included in graph. Similar to previous graph-based clustering methods~\cite{shen2021structure,wang2019linkage,ye2021learning}, data sampling can also provide hard examples which can contribute more to the model training. Therefore, only a fixed number of representative and diverse features are selected for each track, which forms our motivation to conduct a data sampling process as a pre-processing step. Specifically, we first sample $p$ neighbor tracks for each track from its $k$NNs ($k$ nearest neighbors) of face tracks, and then the other two modalities, \emph{i.e.,} body tracks and voice tracks, will be added to the sampled graph according to the association information. If the number of two other modality tracks is less than the given numbers, \emph{i.e.,} $k<p$, the same strategy will be applied to neighbor tracks of pivot track until they reach the given number $p$. After sampling tracks for different modalities, we need to sample a certain amount(refer to $q$ in the main text) of features for each track. Specifically, we set $p=8,q=8$ in our experiment. 

Algorithm~\ref{alg_sampling} shows the details of the density-aware sampling method for sampling features from a track. We denote $s(x_i, x_j)$ as the similarity between two features of the same modality, given a threshold $\tau$, we define a function as $\delta(x_i,x_j)=1$ if $s(x_i,x_j)>\tau$, otherwise $\delta(x_i,x_j)=0$. With the above steps, we can get an initial graph $\mathcal{G}=\{\mathcal{S}_{i}^{\mathcal{F}}\}_{i=1}^p \cup \{\mathcal{S}_{i}^{\mathcal{B}}\}_{i=1}^p \cup \{\mathcal{S}_{i}^{\mathcal{U}}\}_{i=1}^p$ for training and testing, where $\mathcal{S}_{i}^{\mathcal{F}}$, $\mathcal{S}_{i}^{\mathcal{B}}$, $\mathcal{S}_{i}^{\mathcal{U}}$ indicates sampled feature from face tracks, body tracks, and voice tracks, respectively. }
\revise{\subsubsection{Detailed Network Architecture}
 Fig.~\ref{fig_network} shows the detailed network architectures used in RAD-Net. As described in the section of methodology, $\phi_{l,m}$ denotes distribution enhance feature block where $m \in\{\text{face}, \text{body}, \text{voice}\}$, $\sigma_l$ denotes distribution representation similarity block. }
\revise{
\subsubsection{Pseudo-code of Training Procedure}
 Algorithm~\ref{alg_train} and summarize the training procedure of RAD-Net. The entire network of RAD-Net is trained in an end-to-end manner. }
\begin{algorithm}[t]
\small
\caption{One Training Iteration of RAD-Net}
\label{alg_train}
\textbf{Input}: clue features $f(v_i)^0$, initial distribution representations $d(v_i)^0$ in Eq.~\ref{eq_d_init}\\
\textbf{Learnable feature transformation blocks}: distribution representation similarity block $\sigma_l$ and  distribution enhance feature block $\phi_{l,m}$, where $m \in\{\text{face}, \text{body}, \text{voice}\}$ and $l$ indicates $l$-th cycle of the cyclic update;
\begin{algorithmic}[1] 
\FOR{$l$=1 to $L$} 
\STATE Compute the relation of $p^l(v_i=v_j)$ by Eq.~\ref{eq_pij} and Eq.~\ref{eq_pij_dif};
\STATE Update the distribution in a momentum way, \emph{i.e.}, $\bm{d}^l(v_i)\leftarrow\alpha \bm{d}^{l-1}(v_i)+(1-\alpha)\bm{d}^{l}(v_i)$, where $\alpha$ weighs the contribution of the two components;
\STATE Compute the distribution similarity $a^l_{ij}$ by Eq.~\ref{eq_d_similarity};
\STATE Perform feature aggregation to get $\bm{f}^l(v_i)$ by Eq.~\ref{eq_f_aggregation};
\ENDFOR
\STATE Compute $\mathcal{L}\!=\!\lambda_f\mathcal{L}_f+\lambda_d\mathcal{L}_d$ by Eq.~\ref{eq_f_loss} and Eq.~\ref{eq_d_loss};
\STATE Update $\sigma_l$, $\phi_{l,\text{face}}$,$\phi_{l,\text{body}}$, $\phi_{l,\text{voice}}$ by backward propagation.
\end{algorithmic}
\end{algorithm}
\revise{
\subsubsection{Computational Complexity Analysis}
The computational complexity of MAGNET is $O(T \log T + N^2T)$, depending on the graph construction and inference on graphs. Here $T$ is the number of tracks, and $N$ is the number of nodes in the sampled graph. Specifically, we search $k$ nearest neighbors of person tracks to construct the graph, yielding $O(T \log T)$ by Nearest Neighbor (ANN) search algorithm. For inference on graphs, the computation complexity is $O(N^2T)$, considering the pairwise similarity computation for modal fusion. Since $N\!\ll\!T$ in our setting, $O(T \log T\!+\!N^2T)$ is linearithmic to the size of the dataset, which can be easily scaled to large-scale data. Experimentally, clustering on the Buffy dataset (5832 face tracks, 7561 body tracks, 1841 voice tracks) takes 54s on a Tesla T4 GPU.
}

\section{Experiments and Analysis}
\label{sec: experiment}
\subsection{Experimental Setup}

\subsubsection{Datasets}
Our experiments are conducted on a Video Person-Clustering dataset (VPCD)~\cite{brown2021face}. It consists of 32,999 face tracks, 36,724 body tracks, and 9,863 voice tracks, and \revise{features of all modalities are provided for direct use. Generally, the face feature is extracted by SENet-50~\cite{hu2018squeeze} pre-trained on MS-Celeb-1M~\cite{guo2016ms} and fine-tuned on VGGFace2~\cite{cao2018vggface2}, the body feature is extracted by ResNet50~\cite{he2016deep} trained on CSM~\cite{huang2018person}, and the voice feature is extracted by thin-ResNet-34~\cite{xie2019utterance} trained on VoxCeleb2~\cite{chung2018voxceleb2}}. VPCD contains six movies or TV dramas, namely \textit{Hidden Figures}, \textit{About Last Night}, \textit{Sherlock}, \textit{Buffy}, \textit{Friends}, and \textit{TBBT}, respectively, which contain several episodes and many characters. Details about datasets are summarized in Table \ref{tab_sum}.
  \begin{table}[h]
    \centering
    \caption{{Summary of the Datasets}} 
    \resizebox{1\columnwidth}{!}{
    \begin{tabular}{cccccc}
    \toprule[1pt]
    Movies & \textit{Buffy}           & \textit{Friends}       & \textit{Hidden Figure} & \textit{Sherlock}        & \textit{TBBT}          \\ \hline
    \# Characters & 109          & 49          & 24         & 28          & 103          \\ 
    \# Face Tracks & 5832          & 15280          & 1463         & 4902          & 3908          \\
    \# Body Tracks & 7561          & 14447          & 1297         & 4756          & 3756          \\
    \# Voice Tracks$^*$ & 4243          & 13280          & 1509         & 3688          & 2922          \\
    \toprule[1pt]
    \end{tabular}}
    \begin{tablenotes} 
\item \textsuperscript{*}Here we present the original voice tracks. Before using, overlapping pre-processing is needed for screening \cite{brown2021face}. 
\end{tablenotes} 
    \label{tab_sum}
\end{table}

\subsubsection{Evaluation Metrics}
We use Weighted Cluster Purity (WCP), Normalized Mutual Information (NMI)~\cite{ceri2013introduction}, and Character Precision and Recall (CP, CR)~\cite{brown2021face} to evaluate the clustering performance.
Given the predicted cluster set $\Omega=\left\{\omega_{1}, \omega_{2}, \ldots, \omega_{K}\right\}$ and the ground truth cluster set $\mathbb{C}=\left\{c_{1}, c_{2}, \ldots, c_{J}\right\}$, the metrics are calculated by: 
\begin{itemize}
    \item \textbf{Weighted Cluster Purity (WCP)}: WCP computes the purity of a cluster by the number of samples belonging to it. WCP is calculated as $\mathrm{WCP}(\Omega, \mathbb{C})=\frac{1}{\mathrm{~N}} \sum_{k} \max _{j}\left|\omega_{k} \cap c_{j}\right|$. A higher WCP means a better-learned cluster.

    \item \textbf{Normalized Mutual Information (NMI)}~\cite{ceri2013introduction}: NMI measures the trade-off between precision and recall. With $H(\Omega)$ and $H(\mathbb{C})$ as entropies for the predicted cluster set and ground truth cluster set, $I(\Omega, \mathbb{C})$ as the mutual information, the NMI can be calculated as $\operatorname{NMI}(\Omega, \mathbb{C})=\frac{\mathrm{I}(\Omega; \mathbb{C})}{[\mathrm{H}(\Omega)+\mathrm{H}(\mathbb{C})] / 2}$. A higher NMI means a better-learned cluster.
    
    \item \textbf{Character Precision and Recall (CP, CR)}~\cite{brown2021face}: For a cluster, CP is the proportion of tracks that belong to the assigned labels, and CR is the proportion of that label's total ground truth tracks grouped into the cluster.
    In this metric, characters with different numbers of tracks contribute equally. We compute Character Fscore (\textbf{CF}) by $\operatorname{CF}=\frac{2\cdot \operatorname{CP} \cdot \operatorname{CR}}{\operatorname{CP}+\operatorname{CR}}$. A higher CF means better clustering.
    
\end{itemize}
Practically, higher WCP, NMI, CP, and CR indicate better clustering accuracy.

\subsubsection{Cross-Validation}
We use cross-validation to evaluate the clustering performance in VPCD. Specifically, we choose five out of six subsets as the training set and the other one as the testing set. Considering the movie \textit{About Last Night} only has 10 identities, which is not suitable for evaluating our model, we drop this experiment and evaluate the model with the five remaining experiments.


\subsubsection{Implementation Details}\label{datasample}
Adam optimizer \cite{kingma2014adam} is used in all experiments with the initial learning rate as $10^{-3}$ and the learning rate decay as 0.1. The number of generations is set as 2. The loss weights $\lambda_f$ and $\lambda_d$ adjust the relative importance of $\mathcal{L}_f$ and $\mathcal{L}_d$. We set $\lambda_f$ to be 1 and set $\lambda_d$ to be 0.2 for all datasets except for \textit{Friends}. For \textit{Friends}, we set $\lambda_d$ to be 0.3. We demonstrate the selection of $\lambda_f$ and $\lambda_d$ in Table~\ref{tab:sensitive}. We adopt data sampling to build a fixed-size graph to establish mini-batch training and testing for RAD-Net. 

\subsection{Experimental Results}
\begin{table*}[t]
\centering
\caption{{Comparison with baseline methods on VPCD in terms of WCP, NMI, and CF.}}

\begin{tabular}{lccc|ccc|ccc}
\toprule[1pt]
\multirow{2}{*}{Methods}  & \multicolumn{3}{|c|}{\textbf{\textit{TBBT}}}   & \multicolumn{3}{c|}{\textbf{\textit{Buffy}}} & \multicolumn{3}{c}{\textbf{\textit{Sherlock}}} \\ 
 & \multicolumn{1}{|c}{WCP}        & NMI     & CF        & WCP        & NMI      & CF  & WCP        & NMI      & CF \\ 
\midrule
    \emph{Unsupervised methods} \\ \hline
COMIC~\cite{peng2019comic} &\multicolumn{1}{|c}{63.40}  & 56.02  & 52.68  & 71.27  & 49.63  & 53.88  & 59.90  & 30.83  & 42.07 \\ 
AE$^2$-Nets~\cite{zhang2019ae2} &\multicolumn{1}{|c}{67.40}  & 62.50  & 53.19  & 66.61  & 60.32  & 55.00  & 63.19  & 33.95  & 44.35\\
B-ReID~\cite{brown2021face}  & \multicolumn{1}{|c}{80.50}  & 69.70  & 52.16  & 65.00  & 60.90  & 49.58  & 61.20  & 28.90  & 43.95\\ 
COMPLETER~\cite{lin2021completer} &\multicolumn{1}{|c}{59.35}  & 60.88  & 56.85  & 67.81  & 64.48  & 64.33  & 64.82  & 45.53  & 49.44\\ 
B-C1C~\cite{kalogeiton2020constrained}   & \multicolumn{1}{|c}{87.70}  & 69.20  & 44.30  & 73.60  & 58.20  & 37.78  & 77.70  & 41.60  & 35.05\\ 
MuHPC~\cite{brown2021face} & \multicolumn{1}{|c}{\textbf{96.90}}  & 92.80  & \textbf{80.00}  & 85.80  & 76.40  & 67.79  & 84.80  & 60.00  & 56.18\\
\midrule
    \emph{Supervised methods} \\ \hline
MLP & \multicolumn{1}{|c}{87.39}  & 75.05  & 73.41  & 81.41  & 64.88  & 70.08  & 82.37  & 45.37  & 54.48  \\ 
RGCN~\cite{schlichtkrull2018modeling} & \multicolumn{1}{|c}{92.51}  & 82.44  & 81.43  & 91.93  & 67.00  & 75.35  & 86.76  & 51.35  & 65.22\\ 

 

{RAD-Net (our method)} & {96.62} & \textbf{93.12}  & 77.33  & \textbf{90.27}  & \textbf{76.56}  & \textbf{77.33}  & \textbf{93.59}  & \textbf{61.84}  & \textbf{68.74}\\ 
\toprule[1pt]
\multirow{2}{*}{Methods}  & \multicolumn{3}{|c|}{\textbf{\textit{Friends}}}   & \multicolumn{3}{c|}{\textbf{\textit{Hidden Figures}}} & \multicolumn{3}{c}{\textbf{Average\footnotemark[1]}} \\
 & \multicolumn{1}{|c}{WCP}        & NMI     & CF        & WCP        & NMI      & CF  & WCP        & NMI      & CF \\  
\midrule
    \emph{Unsupervised methods} \\ \hline

COMIC~\cite{peng2019comic} &\multicolumn{1}{|c}{63.45}  & 60.91  & 56.27  & 69.08  & 35.90  & 43.27  & 65.42  & 46.66  & 50.00  \\ 
AE$^2$-Nets~\cite{zhang2019ae2} &\multicolumn{1}{|c}{74.60}  & 55.23  & 55.98  & 58.57  & 28.04  & 54.28  & 66.07  & 48.01  & 52.61\\
B-ReID~\cite{brown2021face}  & \multicolumn{1}{|c}{70.90}  & 60.40  & 62.80  & 32.60  & 23.40  & 25.58  & 62.04  & 48.66  & 47.36\\ 
COMPLETER~\cite{lin2021completer} &\multicolumn{1}{|c}{55.60}  & 52.66  & 50.09  & 50.71  & 31.85  & 41.91  & 59.66  & 51.08  & 53.19\\ 
B-C1C~\cite{kalogeiton2020constrained}   & \multicolumn{1}{|c}{85.30}  & 77.10  & 70.14  & 76.20  & 69.80  & 52.64  & 80.10  & 63.18  & 48.32  \\ 
MuHPC~\cite{brown2021face} & \multicolumn{1}{|c}{90.80}  & 83.10  & \textbf{87.15}  & 77.60  & 70.30  & 55.28  & 87.18  & 76.52  & 69.32 \\ 
\midrule
    \emph{Supervised methods} \\ \hline
MLP & \multicolumn{1}{|c}{86.95}  & 71.75  & 78.23  & 91.21  & 61.36  & 44.09  & 85.87  & 63.68  & 65.36  \\ 
RGCN~\cite{schlichtkrull2018modeling} & \multicolumn{1}{|c}{84.85}  & 69.58  & 72.99  & 84.77  & 66.80  & 51.22  & 88.16  & 67.43  & 70.01\\ 

{RAD-Net (our method)} & {\textbf{93.61}}  & \textbf{85.07}  & 86.66  & \textbf{92.28}  & \textbf{78.69}  & \textbf{63.99}  & \textbf{93.27}  & \textbf{79.06}  & \textbf{75.38}\\ 
\toprule[1pt]

\end{tabular}
\begin{tablenotes} 
\item ~~~~~~~~~~~~~~~~~~~~~~~~\textsuperscript{1}The `Average' column is obtained by computing the mean values across the five cross-validations of VPCD.
\end{tablenotes} 
\label{tabcomp}
\end{table*}

\subsubsection{Person Clustering}
We compare our method with two kinds of approaches. One is unsupervised approaches, including COMIC~\cite{peng2019comic}, AE$^2$-Nets~\cite{zhang2019ae2}, B-ReID~\cite{brown2021face}, COMPLETER~\cite{lin2021completer}, B-C1C~\cite{kalogeiton2020constrained}, MuHPC~\cite{brown2021face}. The other is supervised methods, including MLP and RGCN~\cite{schlichtkrull2018modeling}.
The comparison with the state-of-the-art methods is shown in Table~\ref{tabcomp}. Our method outperforms the best benchmark MuHPC by \textbf{6.1\%} on average in WCP, \textbf{2.5\%} in NMI, and \textbf{6\%} in CF, by automatic termination test protocol~\cite{brown2021face} with the unknown number of clusters. 

\revise{Overall, our method can outperform all the benchmarks based on NMI, which is reasonable since NMI is the most related metric for evaluating clustering. Based on WCP and CF metrics, there are two datasets (TBBT and Friends) where we don't have the universal advantage. The reason lies in the different characteristics of the evaluation metrics themselves. (1) WCP weights the purity of a cluster by the number of samples belonging to it; WCP is highest at 1 when within each cluster, all samples are from the same class, so it is not a comprehensive evaluation metric since the model can tend to learn more small repetitive clusters. (2) As for the CF metric, it evaluates algorithm performance at the person level, so it's easily influenced by characters who appear infrequently. For example, TBBT and Friends contain many small casts, so RAD-Net may have lower performance on CF metric compared with other methods.} 

We analyze our method and the most competitive MuHPC in detail. MuHPC utilizes face, body, and voice information to perform person clustering, so it outperforms B-ReID (only with body information) and B-C1C (with face and body information). However, MuHPC designs three different rules manually to utilize all modality information, which may fail to capture the diverse person distributions in the wild. In contrast, our method can capture the complex person distribution by learning the affinities between person clues with distribution representations.
Moreover, RAD-Net improves the most in the dataset with frequent scene switching and long-tail person distributions (\textbf{+1.84\%} NMI on \textit{Sherlock}, \textbf{+8.39\%} NMI on \textit{Hidden Figures} and \textbf{+1.97\%} NMI on \textit{Friends}), which further illustrates the superiority of our RAD-Net in dealing with complex scenarios. 

\textbf{Compared with Multi-view Clustering}: Furthermore, we compare RAD-Net with three multi-view clustering methods, including AE$^2$Nets~\cite{zhang2019ae2}, COMIC~\cite{peng2019comic} and COMPLETER~\cite{lin2021completer}, where different modalities can be treated as different views. These methods fail to keep shared information among different modalities to guarantee information consistency due to weak feature correlations. 

\textbf{Compared with Supervised Methods}: Lastly, we compare RAD-Net with the supervised method MLP and RGCN~\cite{schlichtkrull2018modeling}. With ground truth information, MLP projects clue features of a person into a joint space, whereas RGCN fuses multi-modal information from the same track, which achieves lower performances (-15.38\% NMI and -11.63\% NMI, respectively) than our method. Even with ground truth labels, it is difficult to address the semantic gaps between different modalities in feature space. By contrast, our RAD-Net adopts a relation-aware distribution representation, which is \textbf{modality-agnostic} and friendly for fusing information across different modalities, to avoid these problems.

\begin{table}[t]
    \centering
    \caption{{Person clustering results with noisy associations with noisy ratio from 0 to 0.5. }} 
    {
    \begin{tabular}{cccccc}
    \toprule[1pt]
    $\rho$ & \textit{TBBT}           & \textit{Sherlock}       & \textit{Hidden Figure} & \textit{Friends}        & \textit{Buffy}          \\ \hline
    0             & \textbf{93.11} & \textbf{61.84} & \textbf{78.30} & \textbf{85.07} & \textbf{76.56} \\ 
    0.1           & 91.97          & 61.62          & 78.09         & 84.09          & 75.62          \\
    0.2           & 91.67          & 61.39          & 77.55          & 83.51          & 74.89          \\
    0.3           & 90.31          & 61.58          & 77.42          & 82.78          & 74.76          \\
    0.4           & 89.40          & 60.82          & 77.02          & 81.76          & 73.99          \\
    0.5           & 88.82          & 60.61          & 76.49          & 81.14          & 73.95 \\ 
    \toprule[1pt]
    \end{tabular}} 
    \label{tab_noise}
\end{table}
 
\subsubsection{Person Clustering with Noisy Associations}
Person clustering relies on the given association information across different modalities. In crowd scenes, a person's face may be mistakenly associated with another person's body when two persons stand too close, which brings the noise to inter-modality association information. 
  
To prove the robustness of RAD-Net against noisy associations, we simulate the mistakenly associated body by randomly exchanging the feature between the tracks at a given probability $\rho$, which is denoted as the noisy ratio. It controls the ratio of noisy associations after the random exchange. We set $\rho$ from 0.1 to 0.5. As shown in Table~\ref{tab_noise}, even though the noisy ratio is 0.5, there is only a 4.1\% NMI decrease of our method in \textit{TBBT}. The reason might be that the distribution representation construction and update takes the relation information across all clues, so the distribution representation is robust to data with noisy associations.

\subsection{Ablation Study and Sensitivity Analysis}
\subsubsection{Effectiveness of Feature and Distribution}
To investigate the effectiveness of feature and distribution representation for clues in the graph, we remove them from the original model. (1) \textit{Feature only} is the model without distribution representation, where feature enhancement is performed by feature similarity. Clues with the same modality are clustered separately, and clues with different modalities are grouped by the co-occurrence in the same person track afterward. (2) For \textit{distribution only}, feature representation is removed. Therefore, the intra-modality feature similarities are fixed by the original features. The results are presented in Table~\ref{tab_ablation}. With the comparison between \textit{feature only} and RAD-Net, distribution representation improves the model by 4\% in NMI because it can obtain identity-based information from all modalities to perform clustering; However, feature representation can only get single-modal and pair-wise similarity for clustering. The 3.8\% improvement on NMI of RAD-Net compared to \textit{distribution only} demonstrates that directly adopting the original features can not refine the distribution well, because the enhanced features can aggregate valuable modality-specific information in the feature space.

\begin{table}[t]
    \centering
    \caption{{Ablation study of RAD-Net.}} 
    \begin{tabular}{cccccc}
    \toprule[1pt]
    Method  & WCP   & CP    & CR    & CF    & NMI   \\ \hline
    \textit{feature only} & 92.60 & 89.07 & 60.32 & 71.93 & 75.07 \\ 
    \textit{distribution only} & 92.33 & 91.90 & 62.80 & 74.61 & 75.28 \\ 
    RAD-Net\_f & 95.92  & 91.03   & 63.69  & 74.94    & 75.34 \\
    RAD-Net\_fb & 90.70 & 88.0    & 64.57  & 74.48    & 76.99 \\
    RAD-Net\_fv & 95.28 & 90.66   & 64.87  & \textbf{75.63}    & 75.77 \\
    RAD-Net  & 93.27 & 87.32 & 66.31 & \textbf{75.38} & \textbf{79.06} \\
    \toprule[1pt]
    \end{tabular}
    \label{tab_ablation}
\end{table}

\subsubsection{Effectiveness of Multi-modality Clues}
Table~\ref{tab_ablation} shows that person clustering with clues from all modalities performs better than with clues from partial modalities. RAD-Net\_f, RAD-Net\_fb, and RAD-Net\_fv denote person clustering with face clues only, face+body clues, and face+voice clues, respectively. 
Comparing RAD-Net\_fb, RAD-Net\_fv with RAD-Net\_f, we can see using additional body or voice for clustering can bring improvement by 1.65\% and 0.43\% NMI, respectively. Compared to RAD-Net\_fb and RAD-Net\_fv, our RAD-Net using all three modalities can improve the person clustering by 2.07\% and 3.29\%, respectively. With our relation-aware distribution representation, the benefits of multi-modality are guaranteed and maximized. 


\begin{figure*}[t]
\centering
\includegraphics[width=0.95\textwidth]{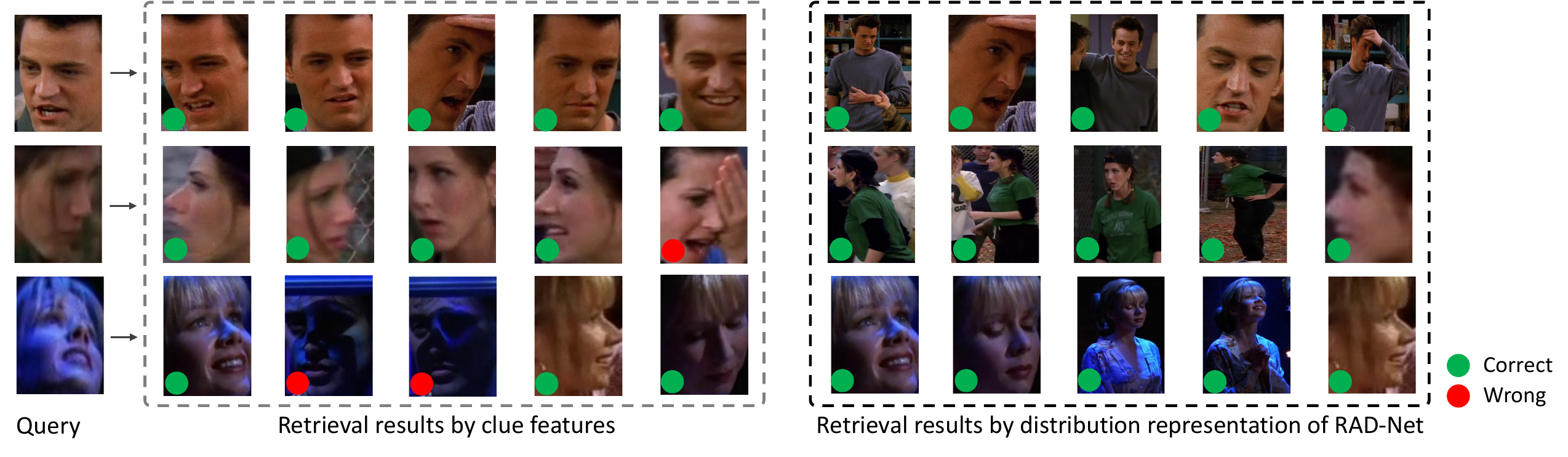} 
\caption{\small{Visualization of retrieval results, with three persons as query (left); Results from feature space (mid) and from RAD-Net distribution space (right). 
}}
\label{fig:retrieval}
\end{figure*}

\begin{table}[h]
\centering
\caption{Sensitivity Analysis of $\eta$, $\lambda_f/\lambda_d$, and Generations, respectively.}
\begin{tabular}{c c|c c|c c}
\toprule[1pt]
$\eta$ & NMI   & $\lambda_f/\lambda_d$ & NMI            & Generations & NMI  \\\hline
0.5    & 76.42 & 0.1                   & 78.36          & 1           & 91.63 \\
0.6    & 78.42 & \textbf{0.2 }                  & \textbf{79.13} & \textbf{2}           & \textbf{93.12} \\
\textbf{0.7}    & \textbf{79.13} & 0.4                   & 78.65          & 3           & 91.61 \\
0.8    & 78.75 & 0.6                   & 78.06          & 4           & 90.91 \\
0.9    & 78.22 & 0.8                   & 78.03          & 5           & 91.17 \\
\toprule[1pt]
\end{tabular}
\label{tab:sensitive}
\end{table}
\subsubsection{Sensitivity of $\eta$}
As shown in Eq.~\ref{eq_d_init}$, \eta$ is the value initialized for $\bm{d}(v_i)_j^0$. We tune the $\eta$ in VPCD and show the average NMIs in Table~\ref{tab:sensitive}. When $\eta=0.5$, the distribution $\bm{d}(v_i)_j$ of samples from the same track and different tracks will be set to equal. In this case, NMI will decrease because of the ignorance of the track information in the dataset. When $\eta$ is high, the NMI will decrease because RAD-Net loses the ability to tolerate noisy data. $\eta=0.7$ is the experimentally optimal choice for soft initialization.

\subsubsection{Sensitivity of $\lambda_f/\lambda_d$}
 The weight $\lambda_f$ and $\lambda_d$ indicates the contribution of $\mathcal{L}_f$ and $\mathcal{L}_d$, respectively. We set $\mu_l^d$ and $\mu_l^f$to be 0.2 when $l<L$, and 1 when $l=L$. We fix $\lambda_d=1$ and then tune $\lambda_f$ from 0.1 to 0.9. As shown in the second column in Table~\ref{tab:sensitive}, the NMI of RAD-Net is low when $\lambda_f/\lambda_d$ is 0.1, because $\mathcal{L}_f$ will provide less contribution to the RAD-Net. When $\lambda_f/\lambda_d>0.2$, the result will be in a downward trend. It is because the distribution graph is more valuable than a feature graph. Excessive weight on the feature graph will affect the optimization of the distribution graph.
 
\subsubsection{Sensitivity of the Number of Generations}
 We investigate the effect of the number of generations in RAD-Net. RAD-Net employs a cyclic update policy to update distribution representations and feature representations. To obtain the effect number of generations of testing results, we report the results conducted on TBBT by the third column in Table~\ref{tab:sensitive}, and two is optimal.

\subsection{Visualization}
\subsubsection{Retrieval of Multi-modal Images} As shown in Fig.~\ref{fig:retrieval}, given a face image, we get the top-5 of the most similar person clues with clue features and distribution representations, respectively. We show that retrieval with the distribution representations is better than clue features in two aspects: (1) Person clues with different modalities can be retrieved by distribution representations but not by clue features. For example, in Fig.~\ref{fig:retrieval}(c), body images could be retrieved given one face image using our distribution representations. This demonstrates our RAD-Net being modality-agnostic. However, using clue features can not achieve this because the clue features are modality-specific, failing to capture cross-modality relations. 
(2) Retrieval using the distribution representations is more robust than using clue features. In Fig.~\ref{fig:retrieval}(b), there are two incorrect samples when using clue features because clues features deteriorate in poor light condition. However, retrieval using distribution representations can avoid this problem because it fuses information from different modalities, making retrieving images with multiple modalities according to identities more robust.

\subsubsection{Visualization of Distribution Similarity} In the supplementary materials (Fig. S3), we also visualize distribution similarities and feature similarities for a sampled graph with only two identities. The visualization illustrates that distribution representations similarities can measure the identity probabilities among different modalities whereas the similarities of clue features can not. 
\begin{table}[t]
\centering
\caption{\small{Multi-view clustering result on VoxCeleb2.}}
{
\begin{tabular}{cccccc}
\toprule[1pt]
Test set& Method  & Precision & Recall & F-score & NMI    \\ \hline
\multirow{6}{*}{\thead{512 \\ identities}} & K-means~\cite{kmeans}  & 79.33   & 63.82 & 70.73 & 90.10  \\ 
& Spetral~\cite{spetral} & 78.70    & 64.04 & 70.62 & 86.81 \\ 
& AHC~\cite{ahc}     & 86.07   & 74.72 & 79.99 & 92.87 \\ 
& ARO~\cite{otto2017clustering}     & 92.26   & 41.41 & 58.09 & 88.13 \\ 
& LGCN~\cite{wang2019linkage}    & 83.70    & 76.83 & 80.12 & 93.12 \\ 
& RAD-Net (\textbf{ours})    & \textbf{92.68}   & \textbf{81.58} & \textbf{86.78} & \textbf{95.51} \\ \hline
\multirow{5}{*}{\thead{2048 \\ identities}} & K-means~\cite{kmeans} & 74.91          & 57.77          & 65.23          & 89.53          \\ 
& AHC~\cite{ahc}    & 81.72          & 67.88          & 74.16          & 92.03          \\
& ARO~\cite{otto2017clustering}    & 35.78          & 44.97          & 39.85          & 50.74          \\ 
& LGCN~\cite{wang2019linkage}   & 81.31          & 68.31          & 74.25          & 92.37          \\ 
& RAD-Net (\textbf{ours})   & \textbf{89.23} & \textbf{76.67} & \textbf{82.48} & \textbf{94.88} \\ 
\toprule[1pt]
\end{tabular}}
\label{tab_4}
\end{table}
\subsection{Results on Multi-view Clustering}
Multi-view clustering is conducted for the instances with multiple features from different views~\cite{yang2018multi,wang2019study}. We extend our method to multi-view clustering by following the same setting as~\cite{wang2019linkage}. We adopt the VoxCeleb2~\cite{chung2018voxceleb2} dataset to evaluate the performance of the multi-view clustering. Similar to~\cite{wang2019linkage}, we split the VoxCeleb2 dataset into a test set with 2048 identities and a disjoint training set with 3434 identities. Also, we sample 512 identities from 2048 identities to get a smaller test set. Several methods, including K-means~\cite{krishna1999genetic}, Spectral~\cite{von2007tutorial}, AHC~\cite{ahc}, ARO~\cite{otto2017clustering}, and LGCN~\cite{wang2019linkage}, are conducted with the test protocol, and the results are presented in Table~\ref{tab_4}. LGCN concatenates face and audio features as joint features and uses GCN to aggregate features for clustering. Our RAD-Net treats different modalities as an instance and adaptively uses the within-modality and inter-modality information to cluster multi-modal person clues in the modal-agnostic distribution space. Our RAD-Net boosts \textbf{6.6\%} and \textbf{8.2\%} F-score on the testing set with 512 and 2048 identities, respectively. 

\section{Conclusion}
\label{sec: conclusion}
This paper aims to cluster a person's multi-modal clues, with different modalities representing rather different and weakly-correlated feature manifolds.
We propose to model multi-modal clues by a relation-aware distribution representation. It employs a graph-based construction mechanism and a cyclic update policy to get a precise distribution representation. Distribution representation is modality-agnostic so that multi-modal clues can be clustered similarly. We demonstrate the effectiveness of our methods on both video person clustering and multi-view clustering datasets.
Our future work plans to extend the proposed model by incorporating more user preference, e.g., user identification information, and uncover its potential in few-shot learning by collaborating with self-supervision-based insights.

\bibliography{main.bib}
\bibliographystyle{IEEEtran}
\end{sloppypar}
\end{document}